\documentclass[runningheads]{llncs}
\usepackage[T1]{fontenc}
\usepackage{graphicx}
\usepackage{hyphenat}
\usepackage{caption, subcaption, float, cuted, amssymb}
\usepackage{amsmath}
\usepackage[dvipsnames]{colortbl,xcolor}
\usepackage[
    colorlinks=true,
    linkcolor=red,
    citecolor=Green,
    urlcolor=RoyalBlue
]{hyperref}
\usepackage{cleveref}
\usepackage{tabularx}
\usepackage{booktabs}
\usepackage{colortbl,xcolor}
\usepackage[accsupp]{axessibility}
\usepackage{xspace}
\usepackage{makecell}
\usepackage{multirow}
\usepackage{comment}
\usepackage{bm}

\usepackage[table]{xcolor}

\newcommand{\suppmat}{\textit{Supp. Mat.}\xspace}
\Crefname{table}{Tables}{Tabs.}

\definecolor{bestgreen}{HTML}{C6EFCE}    
\definecolor{secondgreen}{HTML}{f3f19b}

\definecolor{sttcolor}{HTML}{FCEAA7}
\definecolor{sematicmodulecolor}{HTML}{DAB7D7}
\definecolor{enrichmentcolor}{HTML}{FDE1DF}

\newcommand{\inlineColorbox}[2]{%
  \begingroup
    \setlength{\fboxsep}{1pt}%
    \colorbox{#1}{\hspace*{0.5pt}\vphantom{Ay}#2\hspace*{0.5pt}}%
  \endgroup
}
\newcommand{\stt}{
  \inlineColorbox{sttcolor}{\texttt{STT backbone}}\xspace
}

\newcommand{\semantimodulation}{
  \inlineColorbox{sematicmodulecolor}{\texttt{Semantic Modulation}}\xspace
}
\newcommand{\enrichment}{
  \inlineColorbox{enrichmentcolor}{\texttt{Enrichment Block}}\xspace
}

\definecolor{referencecolor}{HTML}{567BBD}
\definecolor{targetcolor}{HTML}{F384A4}

\newcommand{\reference}{\textbf{\textcolor{referencecolor}{reference motion}}\xspace}
\newcommand{\target}{\textbf{\textcolor{targetcolor}{target motion}}\xspace}
\DeclareRobustCommand{\transition}{%
  \textbf{\textcolor{referencecolor}{tran}}%
  \textbf{\textcolor{referencecolor!50!targetcolor}{siti}}%
  \textbf{\textcolor{targetcolor}{on}}\xspace}

\newcommand{\bestone}[1]{\cellcolor{bestgreen}\text{#1}}

\newcommand{\besttwo}[1]{\cellcolor{secondgreen}\text{#1}}

\usepackage{color}

\begin{document}
\title{Neural Motion Blending Across Arbitrary Character Topologies}
%
%

\author{Luca Cazzola\inst{1}\orcidID{0009-0000-6285-8342}\thanks{Denotes equal contribution.} \and Giulia Martinelli\inst{1,2}\orcidID{0000-0003-3713-3053}$^{\star}$ \and
Nicola Conci\inst{1,2}\orcidID{0000-0002-7858-0928}}
\authorrunning{Cazzola et al.}
%
\institute{University of Trento, Trento 38123, Italy\\
\email{luca.cazzola-1@studenti.unitn.it},
\email{\{giulia.martinelli-2,nicola.conci\}@unitn.it} \and
CNIT, Trento 38123, Italy\\
\url{https://mmlab-cv.github.io/neural_motion_blending/}}

\maketitle              

\begin{abstract}
Motion blending in character animation enables the synthesis of new motions by interpolating between existing examples. Current methods are typically restricted to fixed skeleton topologies, requiring identical or near\hyp{identical} skeletal structures across characters. We present a novel framework for motion blending across heterogeneous skeletons. The proposed architecture combines a semantic encoder, which extracts per\hyp{frame} latent representations of the motion state, with a diffusion-based decoder, which reconstructs character-specific motion conditioned on this latent code. At inference, blended motions are obtained by interpolating the latent representations of two input motions. We train and evaluate the method on the Truebones Zoo dataset using motions defined on both same and distinct skeleton topologies, demonstrating the ability to achieve smooth and plausible blending in a variety of scenarios.

\end{abstract}    
\section{Introduction}
\label{sec:intro}

Animation pipelines involve diverse casts of characters, from humanoids to creatures and stylized avatars, whose skeletal structures vary in topology, proportions, and bone hierarchy. Standard motion blending workflows are generally tied to a fixed skeleton representation, making cross-character interpolation hard to achieve. As a result, animators are often forced to rely on manual correspondence design, retargeting, and cleanup before they can test even simple blended motions~\cite{1604675}. Recent neural approaches have also begun to revisit motion blending as a learnable problem~\cite{Jiang_2025_CVPR,tselepi2025controllable}, aiming to provide more automatic and controllable tools for animators. Still, these approaches do not target cross-topology motion interpolation.

\begin{figure*}[t]
\centering
    \includegraphics[width=\textwidth]{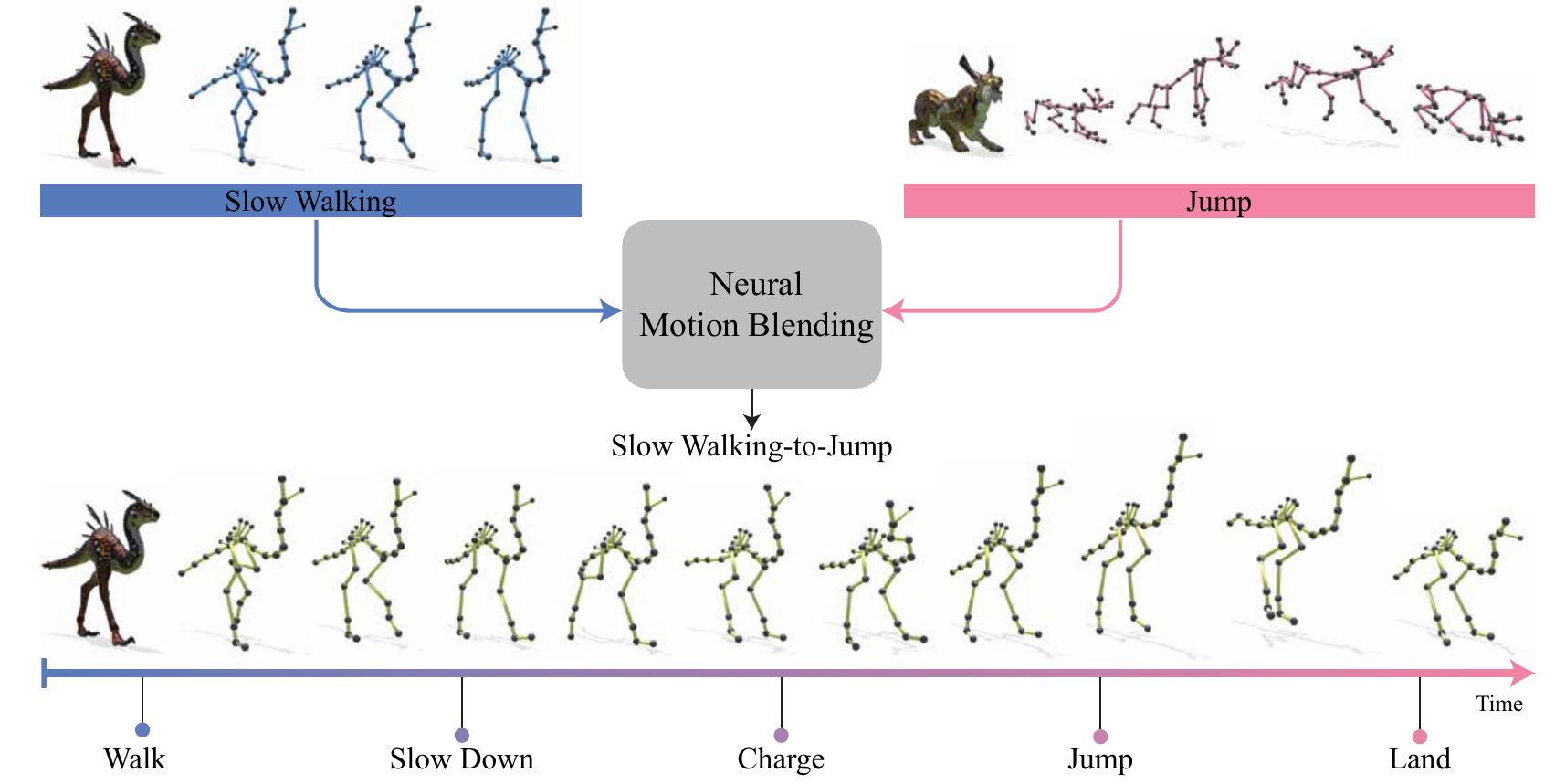}
    \caption{We introduce a unified framework for cross-topology motion blending. Given \reference and a \target, our method generates a \transition while preserving reference skeleton structure. Here, a slow-walking raptor is blended with a jumping lynx using the raptor as reference skeleton, producing a new raptor animation that walks, slows down, charges, jumps, and lands.} 
    \label{fig:teaser}
\end{figure*}

Yet, some progress has been made also in cross-topology motion synthesis~\cite{10.1145/3721238.3730621,wang2025animo} and retargeting~\cite{aberman2020skeleton,10.1145/3757377.3763811,martinelli2025moma}, in part enabled by datasets offering  heterogeneous skeletal structures~\cite{TruebonesZoo2022}. 
However, in synthesis,  motion is generated from scratch, and with retargeting motion is transferred form one character to a second one; thus, neither of them directly addresses the problem of blending two existing motions, especially between different topologies as shown in Figure~\ref{fig:teaser}.

In this work, we introduce a skeleton-agnostic diffusion autoencoding framework for motion blending across heterogeneous characters. Our key idea is to decouple motion semantics from skeleton-specific realization by factorizing each input motion into a semantic code that captures the global motion state, and a stochastic code that captures fine-grained skeleton-specific variation conditioned on it. A graph-based semantic encoder maps motions from arbitrary topologies into a shared latent space, while a diffusion-based decoder reconstructs skeleton-specific motion conditioned on the semantic representation. Blending is then achieved by interpolating semantic codes and composing stochastic components at inference time. Although not the primary goal, the same representation also enables retargeting across heterogeneous skeletons, where we achieve state-of-the-art results against Motion2Motion~\cite{10.1145/3757377.3763811} without requiring explicit joint-level correspondences.

Our contributions can be summarized as follows:
\begin{itemize}
    \item We formalize the task of motion blending across arbitrary skeleton topologies, extending beyond settings that assume identical skeletons.
    \item We introduce a skeleton-agnostic diffusion autoencoding framework that factorizes motion into semantic and stochastic latent components, enabling cross-topology blending.
    \item We demonstrate generalization across heterogeneous skeletons, achieving state-of-the-art performance on the Truebones Zoo dataset~\cite{TruebonesZoo2022}, without requiring explicit joint-level correspondences.
\end{itemize}
\section{Related Work}
\label{sec:related_work}

\textbf{Motion blending.} Motion blending is a core technique in character animation, enabling the synthesis of new motions by interpolating between existing clips in a continuous control space. Classical approaches addressed this problem through motion graphs and statistical formulations of motion interpolation, balancing smoothness, controllability, and efficiency~\cite{Feng2012AnAO,10.1145/566654.566605,10.1145/1073204.1073313}. More recent work has revisited related problems with generative models, including text-conditioned action composition~\cite{Athanasiou2022TEACHTA}, motion editing through blending-based augmentation~\cite{Jiang_2025_CVPR}, seamless transition generation~\cite{barquero2024seamless}, diffusion-based motion priors for transition synthesis~\cite{shafir2024human}. Most directly related to blending as a dedicated task,~\cite{tselepi2025controllable} introduced a single-shot framework for controllable animation blending through temporal conditioning. Despite their differences, these methods operate on fixed-dimensional motion representations tied to a predefined joint set, and therefore assume a shared skeleton topology. By contrast, our work addresses motion blending across characters with heterogeneous skeletal structures, a setting that, to the best of our knowledge, has not been explicitly studied in prior work.

\noindent
\textbf{Cross-topology motion synthesis and retargeting.} Related problems have received increasing attention in motion retargeting and synthesis. Retargeting aims to transfer a motion from a source character to a target character with a different skeletal structure while preserving kinematic plausibility despite differences in hierarchy, proportions, and number of joints. Aberman \textit{et al.}~\cite{aberman2020skeleton} introduced one of the first retargeting methods for heterogeneous skeletons through skeleton-aware convolutions and differentiable pooling. MoMa~\cite{martinelli2025moma} extended this setting with a masked pose autoencoder that supports isomorphic, homeomorphic, and non-homeomorphic skeletons, including characters with additional kinematic chains such as tails. However, these methods are primarily developed and evaluated on humanoid characters, rather than broader cross-species topology variations. More recently, Motion2Motion~\cite{10.1145/3757377.3763811} addressed cross\hyp{topology} motion transfer by formulating retargeting as a conditional patch-based motion matching problem under sparse joint correspondences. Unlike standard retargeting methods, which transfer a source motion to a target skeleton in rest pose, Motion2Motion requires both source and target motion sequences and relies on temporal matching between them. This formulation broadens retargeting beyond the standard transfer setting, but also introduces a dependence on motion-level alignment and sparse structural correspondences between the two skeletons. Recent motion synthesis methods have instead moved toward topology\hyp{agnostic} generation. AnyTop~\cite{10.1145/3721238.3730621} studies motion synthesis conditioned on arbitrary skeletal structures, enabling generation across heterogeneous rigs rather than transfer between matched characters. Despite this progress, it focuses on generating motion for individual skeletons rather than blending between motion sources.

\section{Method}
\label{sec:method}

\subsection{Preliminaries}
\label{sec:preliminaries}
\textbf{Diffusion Autoencoders.}
Diffusion Autoencoders (DiffAE)~\cite{preechakul2021diffusion} augment diffusion models with an explicit semantic representation. Given a clean sample \(x_0\), a semantic encoder extracts a code \(z_{\text{sem}} = E_{\phi}(x_0)\), while the residual stochastic component is represented by \(x_T\), obtained through DDIM inversion~\cite{SongME21}. Each sample is thus represented by the pair \(z=(z_{\text{sem}},x_T)\), where \(z_{\text{sem}}\) captures high-level semantic content and \(x_T\) preserves fine-grained detail. We adopt the same factorized view in the motion domain, with the additional requirement that the semantic representation be shared across heterogeneous skeletal topologies.
\\
\textbf{Motion Diffusion on Arbitrary Skeletons.}
AnyTop~\cite{10.1145/3721238.3730621} is a diffusion model for motion generation across arbitrary skeletal topologies. A skeleton is represented as \(\mathcal{S}=\{\mathcal{P}_{\mathcal{S}},\mathcal{R}_{\mathcal{S}},\mathcal{D}_{\mathcal{S}},\mathcal{N}_{\mathcal{S}}\}\), where \(\mathcal{P}_{\mathcal{S}}\) denotes the rest pose, \(\mathcal{R}_{\mathcal{S}}\) the pairwise joint relations, \(\mathcal{D}_{\mathcal{S}}\) the topological distances, and \(\mathcal{N}_{\mathcal{S}}\) the joint names. A motion sequence is represented as \(X\in\mathbb{R}^{N\times J\times D}\), where \(N\) is the number of frames, \(J\) the number of joints, and \(D=13\) the feature dimension per joint, including root-relative position, 6D rotation~\cite{zhou2019continuity}, linear velocity, and foot-contact label. We use AnyTop as the diffusion backbone for motion decoding.

\subsection{Architecture}
\label{sec:arch}

\begin{figure}[t]
    \centering
    \includegraphics[width=\linewidth]{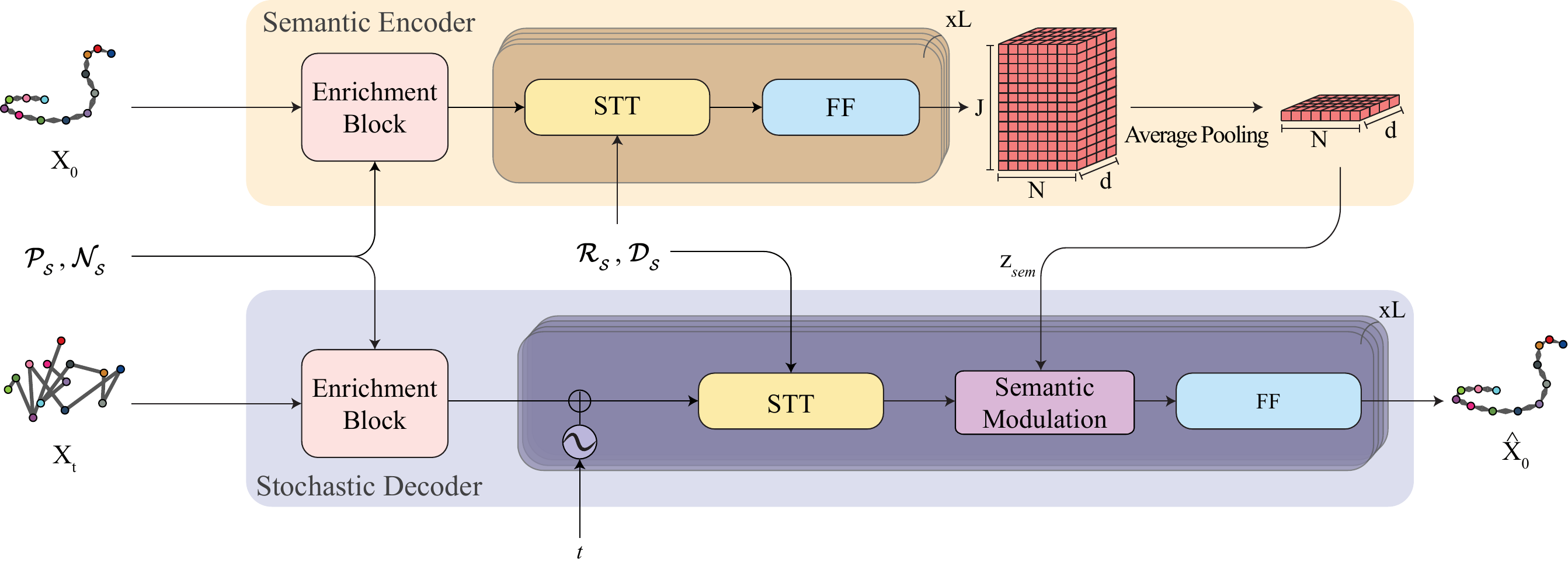}
    \caption{\textbf{The proposed architecture.} Our framework consists of a \textit{Semantic Encoder} and a \textit{Stochastic Decoder}. The former processes the clean motion \(X_0\) through the \enrichment and \stt, then applies average pooling over joints to obtain the semantic code \(z_{\text{sem}}\). The latter takes noisy motion \(X_t\) and predicts clean motion \(\hat{X}_0\), conditioned through the \semantimodulation.}
    \label{fig:arch}
\end{figure}

Fig.~\ref{fig:arch} provides an overview of the proposed architecture, which consists of a \textit{Semantic Encoder} and a \textit{Stochastic Decoder}. We adopt the motion representation introduced in Sec.~\ref{sec:preliminaries} and build on the AnyTop backbone~\cite{10.1145/3721238.3730621}. In particular, we use the same \enrichment to embed the input motion together with skeleton-specific priors, including the rest pose $\mathcal{P}_{\mathcal{S}}$ and the joint descriptions $\mathcal{N}_{\mathcal{S}}$, and the same Spatio-Temporal Transformer \stt, whose layers combine topology-aware spatial attention with local temporal attention through $\mathcal{R}_{\mathcal{S}}$ and $\mathcal{D}_{\mathcal{S}}$. The same \enrichment and \stt are used for both the clean motion $X_0$ in the \textit{Semantic Encoder} and the noisy motion $X_t$ in the \textit{Stochastic Decoder}. We refer the reader to~\cite{10.1145/3721238.3730621} for further details.

\noindent
\textbf{Semantic Encoder.}
Inspired by Diffusion Autoencoders~\cite{preechakul2021diffusion}, we introduce a \textit{Semantic Encoder} to extract a compact semantic representation from the clean motion and use it to condition the stochastic generation process. The encoder processes $X_0$ 
producing a per-frame latent representation $z_{\text{sem}} \in \mathbb{R}^{N \times d}$, where $d$ is the embedding dimension of the\stt. Each latent vector summarizes the global semantic state of the motion at the corresponding frame, including high-level aspects such as motion category, phase, and overall pose dynamics. To enforce a fixed-dimensional representation independently of skeleton topology or joint count, we apply Global Average Pooling (GAP) across the joint dimension at the output of the feed-forward block, yielding a simple topology-agnostic bottleneck.

\noindent
\textbf{Stochastic Decoder.}
Given a noisy motion $X_t$ at timestep $t$, the decoder predicts the corresponding clean motion $\hat{X}_0$ conditioned on the semantic latent $z_{\text{sem}}$. As in the \textit{Semantic Encoder}, the input is first processed 
and conditioning is enabled by the \semantimodulation module inserted after each STT layer. Rather than using standard cross-attention, we adopt a FiLM-style conditioning strategy better matched to our setting, where conditioning is provided by a single vector per frame. With a single key-value token, the attention distribution becomes trivial, and the output reduces to an additive projection of the semantic code without meaningful token-wise interaction. Inspired by FiLM~\cite{perez2018film} and Gated Linear Units~\cite{dauphin2017language}, we instead use a learned modulation mechanism that provides joint-wise multiplicative conditioning. Formally, given a joint feature $h_{n,j}$ and the semantic code $z_{\text{sem}}^{(n)}$ at frame $n$, the modulation is defined as
\begin{equation}
    \tilde{h}_{n,j} = h_{n,j} + g(h_{n,j}, z_{\text{sem}}^{(n)}) \odot \phi(z_{\text{sem}}^{(n)}),
\end{equation}
where $g(\cdot)$ is a learned sigmoid gating function implemented as a two-layer MLP, $\phi(\cdot)$ projects the semantic code to the joint feature space, and $\odot$ denotes element-wise multiplication. Since $g(\cdot)$ depends through concatenation on both $h_{n,j}$ and $z_{\text{sem}}^{(n)}$, each joint is free to modulate semantic information according to its local feature state. This conditioning mechanism allows the decoder to remain controllable when driven by interpolated semantic codes at inference time.

\subsection{Training Objective.}
We adopt the same training objective as AnyTop~\cite{10.1145/3721238.3730621}. Given a clean motion \(X_0\), its noised counterpart \(X_t\), and the target skeleton \(\mathcal{S}\), the model predicts the clean sample \(\hat{X}_0\). Training is based on a reconstruction loss on \(\hat{X}_0\), complemented by a geodesic loss on joint rotations to better reflect distances on the rotation manifold. The final objective is
\begin{equation}
\mathcal{L} = \mathcal{L}_{\text{simple}} + \lambda_{\text{rot}}\mathcal{L}_{\text{rot}},
\label{eq:loss}
\end{equation}
where \(\mathcal{L}_{\text{simple}}\) is an \(\ell_2\) loss on the predicted clean motion and \(\mathcal{L}_{\text{rot}}\) is computed on rotation matrices obtained from the 6D representation via Gram--Schmidt~\cite{zhou2019continuity}. The \textit{Semantic Encoder} is trained jointly with the \textit{Stochastic Decoder} through the same objective. Following DiffAE~\cite{preechakul2021diffusion}, no explicit regularization is applied to $z_{\text{sem}}$; instead, the information bottleneck induced by GAP provides the structural constraint that organizes the latent space, as supported by the interpolation results discussed in Sec.~\ref{sec:results}.

\begin{figure}[t]
    \centering
    \includegraphics[width=\linewidth]{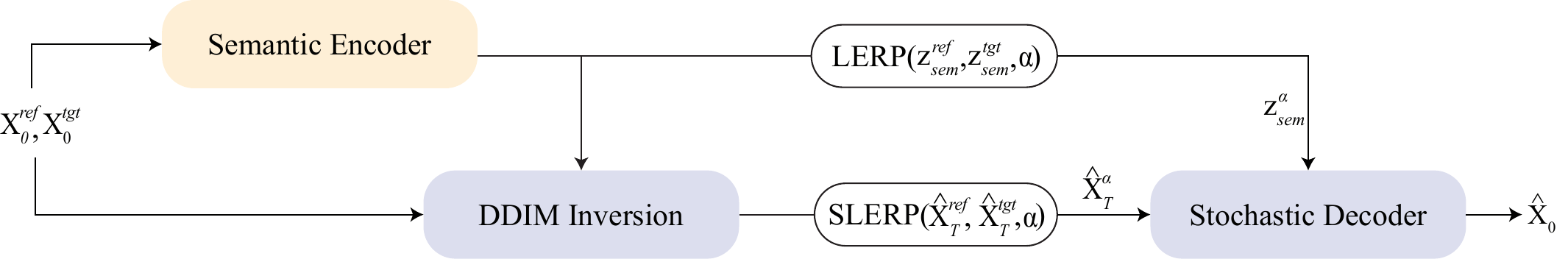}
    \caption{\textbf{Inference: Sampling and Blending.} The reference and target motions $(X^{ref}_0, X^{tgt}_0)$ are encoded into semantic latents and inverted stochastic subcodes. Semantic blending is performed with $\mathrm{LERP}$, while stochastic blending is performed with $\mathrm{SLERP}$. The resulting codes, \(z_{\mathrm{sem}}^{\alpha}\) and \(\hat{X}_T^{\alpha}\), are finally decoded into the blended motion \(\hat{X}_0\).}
    \label{fig:blend_pipeline}
\end{figure}

\subsection{Inference: Sampling and Blending}
\label{sec:blends}

At inference, our method generates motion blends by manipulating the factorized latent code \(z=(z_{\text{sem}}, \hat{X}_T)\), where \(\hat{X}_T\) denotes the stochastic subcode recovered by DDIM inversion. Fig.~\ref{fig:blend_pipeline} illustrates the full pipeline. Given two input motions, a reference motion \(X_0^{\text{ref}}\) and a target motion \(X_0^{\text{tgt}}\), the goal is to synthesize an output motion \(\hat{X}_0\) on the reference skeleton \(\mathcal{S}\) while transitioning from the reference to the target according to a temporal blending schedule \(\alpha(n)\in[0,1]\), where \(n\) indexes the output frames. For readability, in the following we write \(\alpha\) in place of the frame-dependent blending coefficient \(\alpha(n)\).
In the semantic branch, the clean motions are independently encoded by the \textit{Semantic Encoder}, producing semantic latents \(z_{\text{sem}}^{\text{ref}}\) and \(z_{\text{sem}}^{\text{tgt}}\). In the stochastic branch, both motions undergo DDIM inversion~\cite{SongME21}, yielding stochastic subcodes \(\hat{X}_T^{\text{ref}}\) and \(\hat{X}_T^{\text{tgt}}\).
Following~\cite{preechakul2021diffusion}, the semantic and stochastic subcodes are blended through
\begin{equation}
z_{\text{sem}}^{\alpha}=\mathrm{LERP}\!\left(z_{\text{sem}}^{\text{ref}},z_{\text{sem}}^{\text{tgt}},\alpha\right), \qquad
\hat{X}_T^{\alpha}=\mathrm{SLERP}\!\left(\hat{X}_T^{\text{ref}},\hat{X}_T^{\text{tgt}},\alpha\right).
\end{equation}
Applying SLERP to the stochastic subcodes preserves the norm of the interpolated noise, which is expected to lie on the Gaussian shell induced by the diffusion process. In the in-skeleton case, both the semantic and stochastic subcodes can be directly interpolated. In the cross-skeleton case instead $\hat{X}_T^{\text{tgt}}$ cannot be meaningfully reused, as stochastic subcodes are intrinsically tied to their relative skeleton topology. We therefore replace $\hat{X}_T^{\text{tgt}}$ with unit Gaussian noise $\epsilon \in \mathcal{N}(0,I)$ while keeping the semantic interpolation unchanged. This preserves the semantic transition, yet allowing the decoder to synthesize topology-consistent structural details for the output skeleton \(\mathcal{S}\), at the cost of losing fine-grained stochastic details from the target motion.

The blended semantic code and the corresponding stochastic subcode are finally passed to the \textit{Stochastic Decoder}, which synthesizes the output motion \(\hat{X}_0\) on \(\mathcal{S}\) skeleton. When at least one stochastic subcode matches \(\mathcal{S}\) structure, the model relies on DDIM sampling to exploit stochastic subcodes, otherwise, as in cross-skeleton transfer, the model applies Denoising Diffusion Probabilistic Model (DDPM) sampling~\cite{ho2020denoising}. In this case, the stochastic component is randomly sampled rather obtained through DDIM inversion.

\section{Experiments}
\label{sec:results}

\begin{figure}[t]
    \centering
    \includegraphics[width=0.88\linewidth]{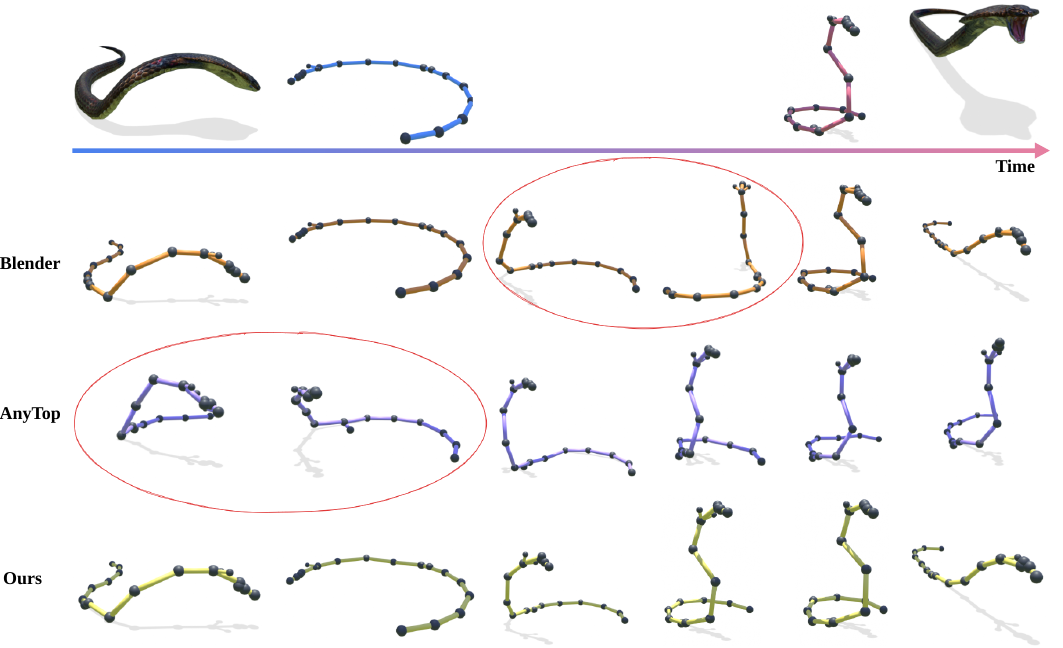}
    \caption{\textbf{Qualitative results: (A) In\hyp{Skeleton} Blending.}}
    \label{fig:inskel}
\end{figure}
\newcolumntype{Y}{>{\centering\arraybackslash}X}

\begin{table}[t]
\centering
\scriptsize
\renewcommand{\arraystretch}{1.25}
\setlength{\tabcolsep}{3pt}
\caption{(A) In\hyp{Skeleton} Blending.}
\label{tab:inskel}
\resizebox{\textwidth}{!}{%
\begin{tabularx}{1.12\textwidth}{@{}lYYYYY@{}}
\toprule
\textbf{Method} &
\mbox{\textbf{FID $\downarrow$}} &
\mbox{\textbf{Proximity $\downarrow$}} &
\mbox{\textbf{Jerk Ratio $\sim 1$}} &
\mbox{\textbf{IntraDivDiff $\downarrow$}}\\
\midrule
Blender NLA~\cite{Blender} & \besttwo{$15.14 \pm 0.60$} & \besttwo{$0.15 \pm 0.15$} & \bestone{$1.64 \pm 1.31$} & \besttwo{$0.11 \pm 0.14$} \\
\midrule
AnyTop (DDIM)~\cite{10.1145/3721238.3730621} & $19.24 \pm 0.45$ & $0.20 \pm 0.12$ & $5.45 \pm 10.09$ & $0.13 \pm 0.09$ \\
Ours (DDIM)   & \bestone{$13.99 \pm 0.32$} & \bestone{$0.13 \pm 0.08$} & \besttwo{$1.83 \pm 1.57$} & \bestone{$0.10 \pm 0.06$}\\
\bottomrule
\end{tabularx}%
}
\renewcommand{\arraystretch}{1}
\end{table}

\subsection{Experimental Setup}
\textbf{Datasets.}
We  evaluate our method on the Truebones Zoo dataset~\cite{TruebonesZoo2022}. Truebones contains 1,219 motion clips (147,178 frames) spanning 70 heterogeneous skeletons, including mammals, birds, insects, dinosaurs, fish, and snakes. Skeletons vary substantially in joint count, connectivity, root definition, scale, and naming convention, making the dataset particularly well suited for evaluating cross-topology motion blending.

\begin{figure}[t]
    \centering
    \includegraphics[width=\linewidth]{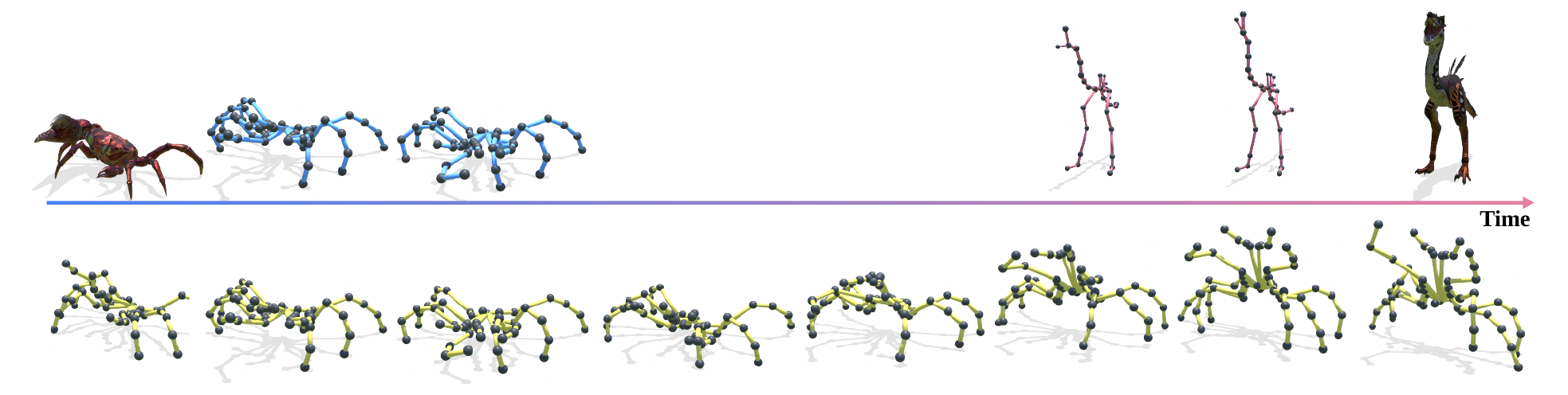}
    \caption{\textbf{Qualitative results: (B) Cross\hyp{Skeleton} Blending.}}
    \label{fig:xskel}
\end{figure}
\newcolumntype{Y}{>{\centering\arraybackslash}X}

\begin{table}[t]
\centering
\scriptsize
\renewcommand{\arraystretch}{1.25}
\setlength{\tabcolsep}{3pt}
\caption{(B) Cross\hyp{Skeleton} Blending.}
\label{tab:xskeleton}
\resizebox{\textwidth}{!}{%
\begin{tabularx}{1.12\textwidth}{@{}lYYYYY@{}}
\toprule
\textbf{Method} &
\mbox{\textbf{FID $\downarrow$}} &
\mbox{\textbf{Proximity $\downarrow$}} &
\mbox{\textbf{Jerk Ratio $\sim 1$}} &
\mbox{\textbf{IntraDivDiff $\downarrow$}} \\
\midrule
Ours (DDPM) &
\besttwo{$17.80 \pm 0.38$} &
\besttwo{$0.22 \pm 0.13$} &
\besttwo{$3.52 \pm 5.63$} &
\besttwo{$0.11 \pm 0.08$}  \\
Ours (DDIM) &
\bestone{$16.81 \pm 0.35$} &
\bestone{$0.20 \pm 0.12$} &
\bestone{$3.51 \pm 5.64$} &
\bestone{$0.10 \pm 0.08$}\\
\bottomrule
\end{tabularx}%
}
\renewcommand{\arraystretch}{1}
\end{table}

\noindent
\textbf{Evaluation Metrics.}
All metrics are computed over the transition region. 
Proximity~\cite{li2022ganimator} and FID~\cite{tselepi2025controllable} assess coherence with the input motions. Proximity operates in the joints space and is therefore restricted to the in-skeleton setting, where comparison against the charachter's real motion pool (GT pool) is well posed. FID is computed in the latent space of the semantic encoder, as no pretrained motion feature extractor generalizes across the heterogeneous topologies considered in our evaluation; to mitigate potential bias, we complement it with joint-space metrics through root relative positions whenever they provide meaningful comparison. \textit{IntraDivDiff}~\cite{10.1145/3721238.3730621} measures whether the internal variability of the transition matches the character's average motion statistics, while \textit{Jerk Ratio} measures smoothness as the ratio between the generated jerk and the mean jerk of the corresponding GT pool, with values close to 1 indicating more common dynamics. In the retargeting setting we report only \textit{FID} and \textit{Jerk Ratio}.

\noindent
\textbf{Baselines.}
We compare against both classical and learning-based baselines. \textbf{Blender}~\cite{Blender} performs geometry-level blending directly in pose space through the Blender Nonlinear Animation editor, without any learned component, and serves as the standard production-level reference. \textbf{AnyTop}~\cite{10.1145/3721238.3730621} is used as a learning-based baseline. We adapt it to the blending setting by applying DDIM inversion to both source motions and interpolating the resulting stochastic codes with SLERP over the transition region. For the retargeting setting on Truebones, we additionally include \textbf{M2M}~\cite{10.1145/3757377.3763811} as a baseline for cross-character motion transfer.

\noindent
\textbf{Setup.}
We evaluate the proposed framework in two main settings: \textbf{(A) In-Skeleton Blending}, where both input motions share the same skeletal topology \(\mathcal{S}\), and \textbf{(B) Cross-Skeleton Blending}, where the two motions originate from characters with different topologies. Unless otherwise specified, we use an \textit{ease-in-ease-out} blending schedule with strength \(1.0\) for \(\alpha(n)\); the effect of this choice is analyzed in Sec.~\ref{sec:ablations}. We further consider the special case \(\alpha(n)=1\), corresponding to \textbf{(C) Motion Retargeting} across different skeletons. Training is performed on the entire Truebones dataset, which lacks inherent ground truths for blending or retargeting tasks. For validation, we randomly pair 5 reference motions per character with 5 target motions. We select motions with at least 40 frames and set a blending region of 30 frames, corresponding to 1.5 seconds.

\subsection{Results}

\begin{figure}[t]
    \centering
    \includegraphics[width=0.88\linewidth]{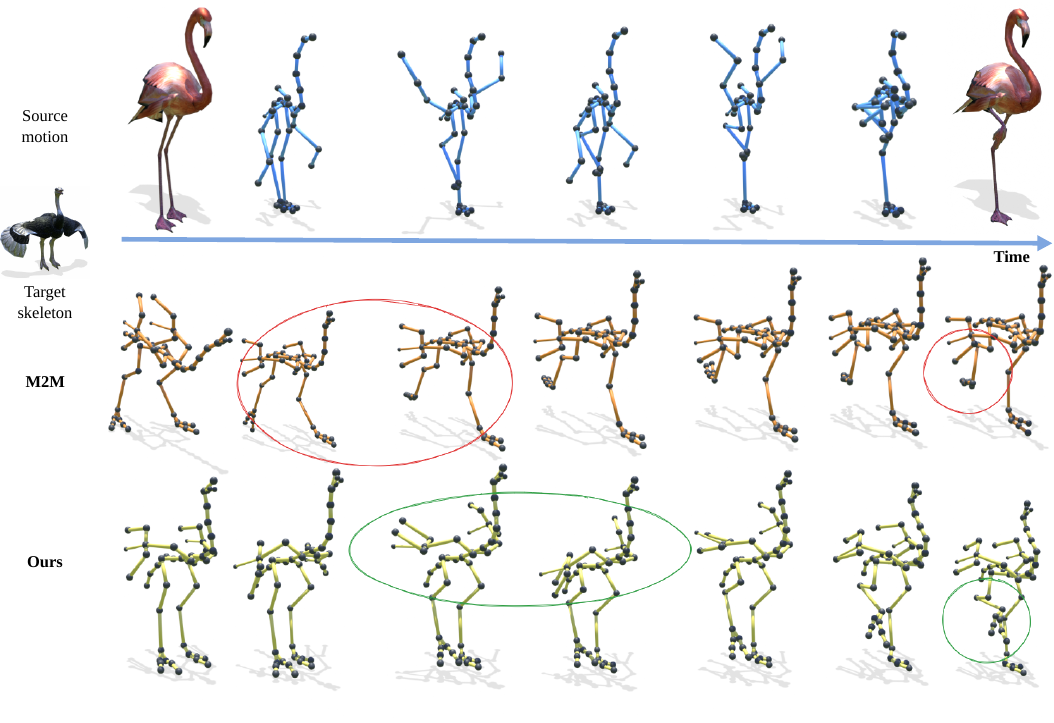}
    \caption{\textbf{Qualitative results: (C) Retargeting.}}
    \label{fig:retargeting}
\end{figure}
\newcolumntype{F}{>{\centering\arraybackslash\hsize=0.8\hsize}X}
\newcolumntype{J}{>{\centering\arraybackslash\hsize=1.2\hsize}X}

\begin{figure}[t]
\centering

\begin{tabular}{@{}m{0.56\textwidth}@{\hspace{0.015\textwidth}}m{0.39\textwidth}@{}}
\captionbox{Retargeting to many.\label{fig:transfer_many}}[0.56\textwidth]{%
    \includegraphics[width=\linewidth]{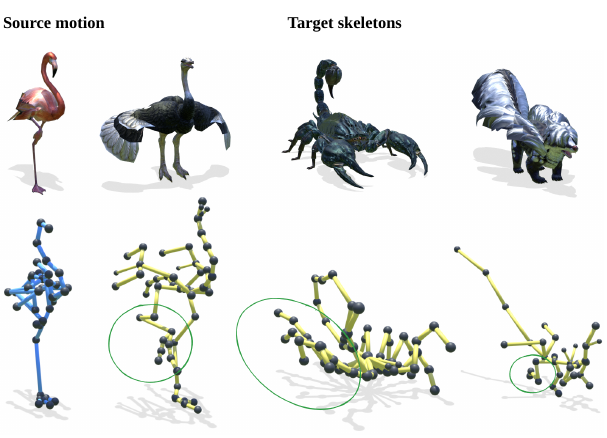}
}
&
\begin{minipage}{0.39\textwidth}
    \centering
    \captionof{table}{(C) Retargeting.}
    \label{tab:retargeting}

    \scriptsize
    \renewcommand{\arraystretch}{1.25}
    \setlength{\tabcolsep}{2pt}
    \resizebox{\linewidth}{!}{%
        \begin{tabular}{@{}lcc@{}}
        \toprule
        \textbf{Method} &
        \textbf{FID $\downarrow$} &
        \textbf{Jerk Ratio $\sim 1$} \\
        \midrule
        M2M~\cite{10.1145/3757377.3763811} &
        \besttwo{$31.31 \pm 0.72$} &
        \bestone{$1.64 \pm 4.11$} \\
        Ours (DDPM) &
        \bestone{$9.64 \pm 0.32$} &
        \besttwo{$2.40 \pm 3.42$} \\
        \bottomrule
        \end{tabular}%
    }
\end{minipage}
\\
\end{tabular}
\end{figure}

\Cref{tab:inskel,tab:xskeleton,tab:retargeting,tab:ablations} shows the \colorbox{bestgreen}{best} and \colorbox{secondgreen}{second best} scores per metric. Qualitative video results are provided in the \suppmat

\noindent
\textbf{(A) In-Skeleton Blending.}
The goal is to generate a blended motion on skeleton \(\mathcal{S}\) from two input motions performed by the same character. Tab.~\ref{tab:inskel} reports the quantitative results. Our model outperforms all baselines in terms of \textit{FID}, \textit{Proximity}, and \textit{IntraDivDiff}, while maintaining a competitive \textit{Jerk Ratio}. AnyTop yields the weakest performance, while Blender produces smoother motions at the cost of poorer transition quality. This trend is visible in Fig.~\ref{fig:inskel}: AnyTop drifts from the reference already at the beginning of the transition, whereas Blender fails to reproduce the correct global reorientation. By contrast, our method produces a gradual and coherent transition that remains faithful.

\noindent
\textbf{(B) Cross-Skeleton Blending.}
In this setting, the two input motions are performed by different characters with different skeletal topologies. Since a direct comparison in the joint space is not applicable across heterogeneous skeletons, and neither Blender NLA nor AnyTop supports cross-skeleton blending, we report results only for our method, using both standard DDIM sampling and stochastic DDPM sampling. The two variants yield comparable results, with DDIM performing slightly better overall, while DDPM highlights the stochastic potential of the framework by generating multiple plausible alternatives from the same input pair. Qualitative results are shown in Fig.~\ref{fig:xskel}, with additional video examples provided in the \suppmat The generated motion remains faithful to the reference at the beginning of the sequence, then progressively shifts toward the target behavior while respecting the reference topology.

\noindent
\textbf{(C) Retargeting.}
When setting constant \(\alpha(n)=1\) our framework reduces to motion retargeting. Since the Truebones dataset does not provide paired motions, we evaluate this setting using only \textit{FID} and \textit{Jerk Ratio}. We additionally compare against M2M~\cite{10.1145/3757377.3763811}, which is the only existing baseline in our comparison specifically designed for retargeting. Since M2M relies on sparse correspondences and motion matching, Table~\ref{tab:retargeting} compares only walking motions with manually aligned leg correspondences. Tab.~\ref{tab:retargeting} shows that our model outperforms M2M by a significant margin in \textit{FID}, while maintaining a competitive \textit{Jerk Ratio}. M2M relies on sparse joint correspondences and semantically similar motion pairs. Our method achieves stronger results without requiring either. Moreover, although \textit{FID} is computed in the latent space of our encoder for both methods, the qualitative comparison in Fig.~\ref{fig:retargeting} is consistent with the quantitative results: M2M fails to reproduce target semantics faithfully, whereas our output remains both faithful to target motion and structurally plausible on reference topology. Additional qualitative results across different characters are presented in Fig.~\ref{fig:transfer_many}, demonstrating consistent retargeting across topologies with varying joint counts and limb structures.

\subsection{Ablation Studies}
\label{sec:ablations}
Tab.~\ref{tab:ablations} reports two ablations in the in-skeleton setting. First, removing rotation (\textit{w/o Rot}) or velocity (\textit{w/o Vel}) features degrades performance on nearly all metrics, showing that both cues are important for stable and coherent blending. Second, we compare different blending schedules \(\alpha(n)\), namely Linear interpolation and Ease interpolation with two strengths. The results are largely comparable, with the main variation appearing in \textit{Jerk Ratio}: stronger easing introduces larger velocity changes and therefore slightly less smooth transitions. We use Ease with strength \(1.0\) in the final model, as it provides the best trade-off between preserving motion dynamics and maintaining smooth transitions.
\newcolumntype{Y}{>{\centering\arraybackslash}X}

\begin{table}[t]
\centering
\scriptsize
\renewcommand{\arraystretch}{1.25}
\setlength{\tabcolsep}{3pt}
\caption{Ablations Studies.}
\label{tab:ablations}
\begin{tabularx}{\textwidth}{@{}llYYYY@{}}
\toprule
\textbf{Ablation} &
\textbf{Setting} &
\mbox{\textbf{FID $\downarrow$}} &
\mbox{\textbf{Proximity $\downarrow$}} &
\mbox{\textbf{Jerk Ratio $\sim 1$}} &
\mbox{\textbf{IntraDivDiff $\downarrow$}} \\
\midrule
\multirow{3}{*}{\centering In-Feats} &
w/o Rot &
$10.72 \pm 0.65$ &
\besttwo{$0.13 \pm 0.06$} &
$2.59 \pm 4.34$ &
\bestone{$0.11 \pm 0.07$} \\
&
w/o Vel &
\besttwo{$10.63 \pm 0.66$} &
$0.13 \pm 0.07$ &
\besttwo{$1.81 \pm 1.28$} &
\bestone{$0.11 \pm 0.07$} \\
&
Full &
\bestone{$9.75 \pm 0.62$} &
\bestone{$0.12 \pm 0.06$} &
\bestone{$1.65 \pm 0.88$} &
\bestone{$0.11 \pm 0.07$} \\
\midrule
\multirow{3}{*}{\centering $\alpha(n)$} &
Linear &
\besttwo{$9.76 \pm 0.60$} &
\bestone{$0.12 \pm 0.06$} &
\bestone{$1.48 \pm 0.71$} &
\besttwo{$0.11 \pm 0.07$} \\
&
Ours (Ease $1.0$) &
\bestone{$9.75 \pm 0.62$} &
\bestone{$0.12 \pm 0.06$} &
\besttwo{$1.65 \pm 0.88$} &
\besttwo{$0.11 \pm 0.07$} \\
&
Ease $2.0$ &
$9.78 \pm 0.63$ &
\bestone{$0.12 \pm 0.06$} &
$1.80 \pm 1.01$ &
\bestone{$0.10 \pm 0.07$} \\
\bottomrule
\end{tabularx}
\renewcommand{\arraystretch}{1}
\end{table}

\section{Conclusions}
\label{sec:conclusion}

We presented a unified framework for motion blending across arbitrary skeletal topologies. By factorizing motion into a shared semantic code and a skeleton-specific stochastic component, our method enables blending between motions performed by different characters without requiring joint correspondences. Experiments on the Truebones Zoo dataset show remarkable performance in both in-skeleton and cross-skeleton settings, while the same representation also supports motion retargeting. We believe this work opens to a new direction for motion authoring tools that are less constrained by fixed skeleton representations and better aligned with the diversity of production animation pipelines.

\noindent
\textbf{Limitations.} In the cross-skeleton and retargeting settings, the target stochastic code is replaced with Gaussian noise, which discards part of available target motion information. Additionally, transitions between semantically distant motions may exhibit artifacts, as the interpolated semantic trajectory must cover a large latent distance without explicit guidance on intermediate poses. Incorporating transition planning or intermediate pose conditioning could help resolve these cases and improve overall blending quality.

%
%
%

%
%
%
\bibliographystyle{splncs04}
\begingroup\sloppy
\bibliography{main}
\endgroup

\clearpage
\setcounter{page}{1}
\appendix
\vspace*{2mm}
\section*{\texorpdfstring{\centering }{}
Supplementary Material}
\vspace*{10mm}
\renewcommand{\thesection}{\Alph{section}}  
\setcounter{section}{0}

In this supplementary material, we first provide implementation details (Sec.~\ref{sec:implementation}), followed by a more in-depth description of the evaluation metrics (Sec.~\ref{sec:metrics}) used in the paper. Finally, we present additional results (Sec.~\ref{sec:additional_results}) and analysis (Sec.~\ref{sec:latent_space}). The supplementary material also includes qualitative results in video format, accessible at \url{https://mmlab-cv.github.io/neural_motion_blending/}. The webpage is organized according to the three experimental settings discussed in the paper: in-skeleton blending, cross-skeleton blending, and retargeting. Each section presents representative video examples and, when applicable, side-by-side comparisons with Blender, AnyTop, and Motion2Motion.

\section{Implementation details}
~\label{sec:implementation}
\noindent
The \textit{Semantic Encoder} and \textit{Stochastic Decoder} employ a symmetrical configuration consisting of $L=4$ blocks with a latent dimensionality of $H=128$. The \enrichment projects the $D=13$ joint\hyp{wise} features into the $H$-dimensional space. 
The diffusion process is set with $T=100$ timesteps, which are integrated into the stochastic decoder using sinusoidal embeddings. Model training is driven by Eq.~\ref{eq:loss} with $\lambda_{rot} = 1$. We train for 500k steps on a single NVIDIA RTX 3090 GPU (24 GB VRAM) using the AdamW optimizer. Hyperparameters include a batch size of $10$ and an initial learning rate of $1 \times 10^{-4}$. We utilize a StepLR scheduler that reduces the learning rate every 10k steps.

We preprocess the Truebones Zoo \cite{TruebonesZoo2022} dataset following the pipeline described in \cite{10.1145/3721238.3730621}. Each joint for every character is represented as described at Sec.~\ref{sec:preliminaries}. To ensure consistency across diverse topologies, we apply character-level Z-score normalization by computing the mean and standard deviation for each specific entity. Skeletons are rotation-normalized to face the $+Z$ direction using the hips and shoulder joints (or their anatomical equivalents) as reference and are grounded by setting the feet at height $Y=0$. Motion clips are restricted to a maximum length of 200 frames; sequences exceeding this threshold are subdivided into multiple segments. Training is performed on clips of length 40 sampled randomly at each epoch.

\section{Evaluation Metrics}
\label{sec:metrics}
Evaluation relies on deep features for FID and root-relative positions for all other metrics. Calculations are performed on the transition zone, which corresponds to the entire target clip in the specific case of motion retargeting.

\newcolumntype{Y}{>{\centering\arraybackslash}X}

\begin{table}[t]
\centering
\scriptsize
\renewcommand{\arraystretch}{1.25}
\setlength{\tabcolsep}{3pt}
\caption{Runtime analysis of the blending pipeline.}
\label{tab:runtime}
\resizebox{0.95\textwidth}{!}{%
\begin{tabularx}{\textwidth}{@{}lYYY@{}}
\toprule
\textbf{Components} &
\mbox{\textbf{Overall $\downarrow$}} &
\mbox{\textbf{Seconds / Joint $\downarrow$}} &
\mbox{\textbf{Seconds / Frame $\downarrow$}} \\
\midrule
Sampling       & $2.487 \pm 0.148$  & $0.051 \pm 0.003$ & $0.022 \pm 0.001$ \\
DDIM Inversion & $4.625 \pm 0.289$  & $0.096 \pm 0.006$ & $0.041 \pm 0.002$ \\\rowcolor{black!6}
IK             & $13.655 \pm 1.483$ & $0.283 \pm 0.031$ & $0.121 \pm 0.013$ \\
\midrule

Full pipeline  & $20.767 \pm 1.602$ & $0.430 \pm 0.033$ & $0.185 \pm 0.014$ \\
\bottomrule
\end{tabularx}%
}
\renewcommand{\arraystretch}{1}
\end{table}

\noindent \textbf{FID (Fréchet Inception Distance).} This metric assesses generative fidelity by comparing the synthesized motion distribution against the ground truth. Since no pretrained topology\hyp{agnostic} motion feature extractor exists, we use the \textit{Semantic Encoder} as feature space and complement FID with joint-space metrics where applicable. 

\noindent \textbf{Proximity.} Consists in the minimum Patch Nearest Neighbour (PNN) distance, derived from GANimator\cite{li2022ganimator}, this metric measures the minimum distance between the synthesized transition and the source animations. In the context of blending, it is interpreted as a "lower-is-better" score where smaller values indicate higher fidelity to the reference and target motions. It serves as a joint-space complement to FID, ensuring the synthesized results remain highly correlated with the control sequences.

\noindent \textbf{Jerk Ratio.} This measures motion smoothness by computing the third-order derivative of joint positions with respect to time. The ratio is derived by normalizing the synthesized jerk against the character's average ground truth jerk, where a value of 1.0 represents average dynamics. It provides critical insight into motion artifacts, such as high-frequency flickering or unnatural jitter, that simple positional metrics might overlook.

\noindent \textbf{IntraDivDiff (Intra-Diversity Ground Truth Difference).} Adopted from \cite{10.1145/3721238.3730621}, this metric measures the absolute difference between the internal statistical diversity of the generated motion and the ground truth set. A value of 0 indicates that the generated transition possesses internal variability identical to the character's average motion statistics. Higher values suggest a deviation from natural movement, making lower scores preferable for maintaining character realism.

\section{Additional Results}
\label{sec:additional_results}

\noindent
\textbf{Runtime Analysis}
Tab.~\ref{tab:runtime} reports the runtime of the main components of our pipeline. The largest computational cost is due to the inverse kinematics (IK) post\hyp{process}, which dominates the overall runtime. In comparison, the learned stages are relatively efficient: sampling requires approximately \(2.5\) seconds, and DDIM inversion approximately \(4.6\) seconds. The complete pipeline takes about \(21\) seconds per blend, with per\hyp{joint} and per\hyp{frame} costs scaling moderately with the size of the skeleton and the sequence length. While this does not enable real-time interaction, it is well suited to offline animation authoring, where the runtime remains negligible compared to the manual effort typically required to an animator to set up correspondences, retarget motions, and clean up a blended transition.

\begin{figure}[t]
    \centering
    \includegraphics[width=\linewidth]{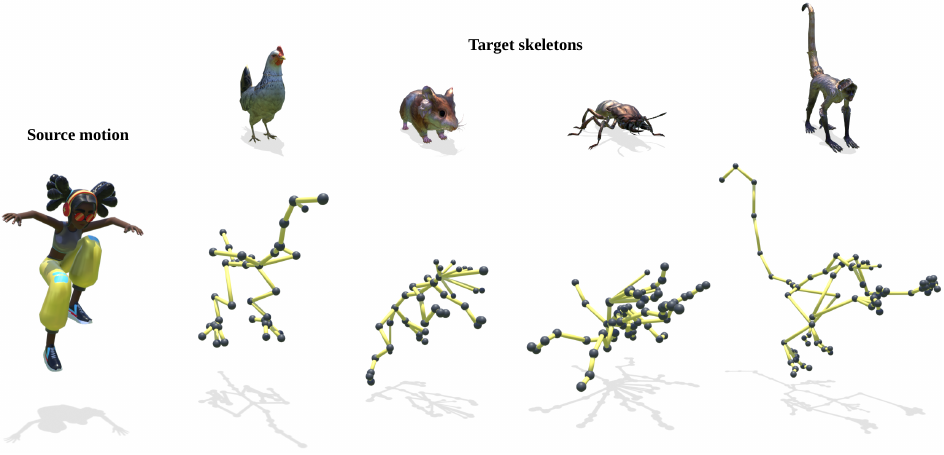}
    \caption{Zero\hyp{Shot} Retargeting from Unseen Topologies}
    \label{fig:cross_dataset}
\end{figure}

\noindent
\textbf{Cross\hyp{Dataset} Retargeting.}
Fig.~\ref{fig:cross_dataset} illustrates an additional generalization experiment in which the source motion originates from a humanoid character. Since training is performed exclusively on the Truebones Zoo dataset, the human skeleton is entirely unseen during training. Nevertheless, the model successfully transfers the motion to morphologically diverse target characters, including species with fundamentally different limb structures and joint counts. This result suggests that the learned semantic space captures motion patterns at a level of abstraction that is not tied to specific skeletal configurations encountered during training.

\section{Latent Space Analysis}
\label{sec:latent_space}

To evaluate if the semantic encoder structures its latent space towards a meaningfull topology\hyp{agnostic} space, we perform a $k$-nearest-neighbour ($k$-NN) classification probe. All motions from the Truebones Zoo dataset are encoded, and their per-frame latents are averaged into a single descriptor. We evaluate a $k$-NN classifier at $k \in \{1, 5, 15, 30\}$ across two downstream tasks: (i) skeleton category (\texttt{flying}, \texttt{snakes}, \texttt{quadrupeds}, \texttt{millipeds}, \texttt{bipeds}) and (ii) action semantics (\texttt{attack}, \texttt{idle}, \texttt{run}, \texttt{die}, \texttt{walk}, \texttt{other}). These labels are derived manually from filenames, making them highly noisy due to naming inconsistencies and the dataset's nature. Crucially, they are used strictly at evaluation time.

\begin{figure}[t]
    \centering
    \includegraphics[width=\linewidth]{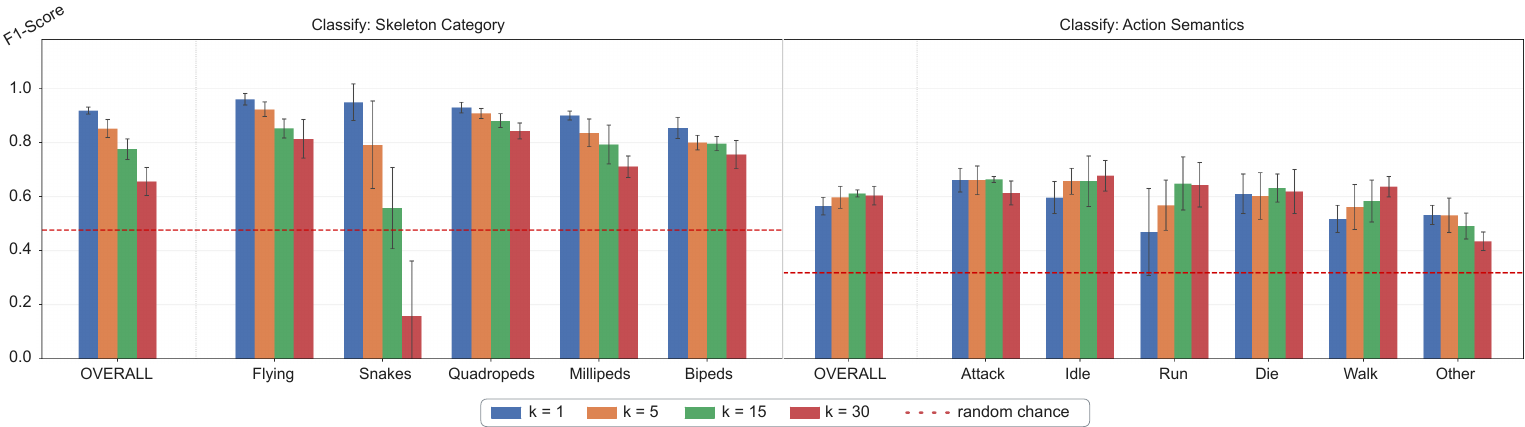}
    \caption{\textbf{k-NN classification of the semantic latent space.} F1 scores at $k \in \{1, 5, 15, 30\}$ for skeleton category (\textit{left}) and action semantics (\textit{right}). The dashed red line indicates the random-chance baseline.}
    \label{fig:knn_ablation}
\end{figure}

As shown in Figure~\ref{fig:knn_ablation}, the encoder implicitly organizes the latent space along morphological and behavioral axes, with all F1-Score probes performing statistically well above the random-chance baseline. This baseline is defined as $\max(1/C, \text{majority-class rate})$, where $C$ is the class count. For skeleton categories (left), the F1-Score at $k \in \{5, 15\}$ reaches $\sim{0.80}$ against a $\sim{0.48}$ baseline, degrading gracefully except for the underrepresented \texttt{snakes} category. For action semantics (right), the F1-Score reaches $\sim0.58$ against a $\sim{0.31}$ baseline. This result is highly significant given the severe label noise and high dataset complexity, where each character averages 20--30 motion samples spanning very diverse actions. Consequently, distinct behaviors like \texttt{attack} and \texttt{idle} are highly reliable, while heterogeneous classes like \texttt{run} and \texttt{other} show more confusion. This structure emerges naturally because the reconstruction objective forces the encoder to capture both body morphology and behavior.

\begin{figure}[t]
    \centering
    \includegraphics[width=\linewidth]{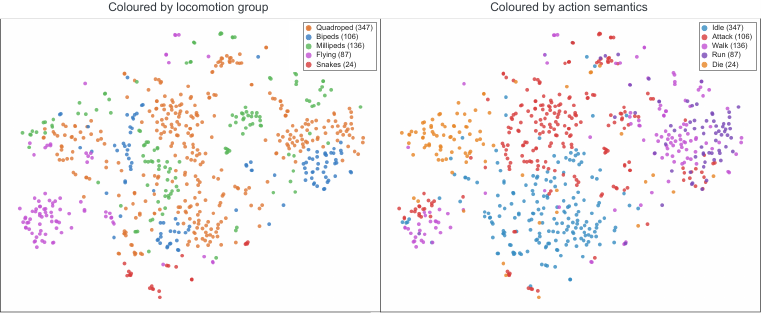}
    \caption{\textbf{t-SNE visualization of the semantic latent space.} The embedding is colored by locomotion group (\textit{left}) and action semantics (\textit{right}), demonstrating local semantic coherence across both dimensions.}
    \label{fig:tsne_ablation}
\end{figure}

Figure~\ref{fig:tsne_ablation} provides a qualitative t-SNE projection of these descriptors. Colored by locomotion group (left) or action semantics (right), the embedding exhibits emerging local neighborhood without global clustering. This simultaneous encoding ensures that semantically similar motions align across different skeleton types, providing the latent structure necessary to support cross-topology blending.

\end{document}